# Benchmarking Meta-heuristic Optimization

**Mona Nasr**
Faculty of Computer and Artificial Intelligence
Department of Information Systems
Helwan University - Cairo, Egypt
m.nasr@helwan.edu.eg

**Omar Farouk**
Faculty of Computer and Artificial Intelligence
Department of Computer Science
Helwan University - Cairo, Egypt
omar_20160269@fci.helwan.edu.eg

**Ahmed Mohamedeen**
Faculty of Computer and Artificial Intelligence
Department of Computer Science
Helwan University - Cairo, Egypt
ahmed.2m@fci.helwan.edu.eg

**Ali Elrafie**
Faculty of Computer and Artificial Intelligence
Department of Computer Science
Helwan University - Cairo, Egypt
ali_20160249@fci.helwan.edu.eg

**Marwan Bedeir**
Faculty of Computer and Artificial Intelligence
Department of Computer Science
Helwan University - Cairo, Egypt
marwan_20160411@fci.helwan.edu.eg

**Ali Khaled**
Faculty of Computer and Artificial Intelligence
Department of Computer Science
Helwan University - Cairo, Egypt
ali_20160245@fci.helwan.edu.eg

-------------------------------------------------------------------**ABSTRACT**-------------------------------------------------------------------
**Solving an optimization task in any domain is a very challenging problem, especially when dealing with nonlinear problems and non-convex functions. Many meta-heuristic algorithms are very efficient when solving nonlinear functions. A meta-heuristic algorithm is a problem-independent technique that can be applied to a broad range of problems. In this experiment, some of the evolutionary algorithms will be tested, evaluated, and compared with each other. We will go through the Genetic Algorithm, Differential Evolution, Particle Swarm Optimization Algorithm, Grey Wolf Optimizer, and Simulated Annealing. They will be evaluated against the performance from many points of view like how the algorithm performs throughout generations and how the algorithm's result is close to the optimal result. Other points of evaluation are discussed in depth in later sections.**

## I. INTRODUCTION

Meta-heuristic Optimization Algorithms, a wide variety of classes fall into the meta-heuristic category, for example, *Evolutionary Algorithms*, *Swarm Intelligence*, and *Stochastic Algorithms*, they are very useful to find a near-optimal solution to nonlinear problems or non-convex functions, of course finding the optimal solution is more efficient but a very expensive task, computation power-wise. However, a near-optimal solution is satisfying. There are a lot of evolutionary algorithms, each one has its technique to find the optimal solution. In this experiment, we will compare five evolutionary optimization algorithms for the same objective function, size of the population, and the number of iterations.

## II. OPTIMIZATION
*A. Local optimization*

Local optimization is the search for the smallest objective value in what is considered a feasible neighborhood. Searching for the best solution or near-optimal solution will require navigating a huge space of possibilities iteratively. [1] But in some problems such as *Traveling Salesman Problem* the search is limited to the space of selected candidate solutions based on some initial computations. But some problems like non-convex problems the usage of local optimization techniques is normally not sufficient to solve such problems. The search is affected heavily by the initial point and doesn't guarantee global optimal.

*B. Global optimization*

Recently more complicated methods are focused on Global Optimization, which is searching for the smallest objective value in all the feasible neighborhoods. There is a big variety of global optimization methods are designed and there are many years to come to introduce even more advanced techniques or methods. The first mention of global optimization was brought up in 1975 [2]. Now decades later the optimization problems have seen some maturity and some of the methods purposed are best when used against some of the problems. Thus, in this experiment, we will be comparing a number of different methods.

## III. LITERATURE REVIEW
*A. Genetic Algorithm*

Computer simulation of evolution was an idea that was put in practice in 1954 by Barricelli, just four years after Alan Turing proposed a learning capable machine. [3] Genetic algorithm (GA) the name itself comes from the fact it's mimicking evolutionary biology techniques. The early adopters of genetic algorithm technique were the biologists and over the next decades, GA use was expanded to many fields, solving many problems with success leading to it being still used in many areas today. The algorithm is built on the concept of evolution where the first generation of the population is evolved into a more fit generation with better genome (features).

Starting with a randomly initialized first generation that is tested against the objective task to keep track who are the fittest individuals, then the subsequent generations are created keeping in mind the best genes/features found in the parent generation. There are many implementations of the genetic algorithm, In this experiment, we'll use (Simple Genetic Algorithm Figure 1) which has three main operations [4] applied to the parents to produce offspring individuals.

Algorithm 1 shows a pseudo code of the Simple GA for minimizing the objective.

---

Algorithm 1 Genetic Algorithm

1: *Population* ← InitializePopulation($Pop_{size}$)
2: $S_{best}$ ← GetBestSolution(*Population*)
3: while *genNum*6= 0 do
4:     *Parents* ← SelectParents(*Population*,$Pop_{size}$)
5:     *Children* ← $\varphi$
6:     for each $P_1,P_2$ in *Parents* do
7:         $C_1,C_2$ ← Crossover($P_1,P_2,Probability_{cx}$)
8:             *Children* ← Mutate($C_1,Probability_{mutation}$)
9:             *Children* ← Mutate($C_2,Probability_{mutation}$)
10:     EvaluatePopulation(*Children*);   11: $S_{best}$ ← GetBestSolution(*Children*)
12:     *Population* ← *Children*
13:     *genNum* ← *genNum*- 1
14: return $S_{best}$

---

The simple three operations of the simple GA are further explained below:

- Selection: It defines the search space before going any further. There are many selection mechanisms used like *tournament, truncate and roulette wheel* we're using the simple tournament method which randomly selects defined number of individuals and then performs a tournament amongst them, choosing the best one and repeating the steps until the offspring generation is formed.
- Mutation: Mutation operations involve direct modification of the genes with some probability of occurring. There are many ways of doing mutations, we'll be using uniform mutation.
- Crossover: Crossover is a genetic operator that mates two individuals (parents) to create a new one. The general idea behind it; Is trying to create a better individual if it contained the best features from both of the parents. We will use in this experiment single point crossover which occurs at a random point with portability of happening.

*B. Particle Swarm Optimization Algorithm*

The particle swarm optimization (PSO) was originally presented by Kennedy and Eberhart [5, 6, 7]. It is inspired by the social behavior of swarms in nature, Every particle of this swarm is searching through n-dimensional space for the best solution. Every particle remembers the best solution he found and also knows the global best solution found by swarm till now. The algorithm is done by repeating fixed steps for several iterations and in each one every particle changes his position according to the following formula:

$$\vec{X}_{id} = \vec{X}_{id} + \vec{V}_{id} \quad \text{(Eq. 3.1)}$$

Where $X_{id}$ is the current position and $V_{id}$ is the current velocity, but first, we need to calculate its velocity according to this formula:

$$\vec{V}_{id} = \omega \vec{X}_{id} + c_1 r_1(\vec{P}_{id} - \vec{X}_{id}) + c_2 r_2(\vec{G} - \vec{X}_{id}) \quad \text{(Eq. 3.2)}$$

Where $\omega$ is the inertia factor. $P_{id}$ is the best solution found by this particle. $G$ is the global best solution found by swarm until now. $c_1$ and $c_2$ are the weights determining the effect of $P_{id}$ and $G$. $r_1$ and $r_2$ are random factors.

As we can see from Eq. 3.2 there are three directions that take part in determining the new position of the particle. That will be more clearly in Figure 2

There are two opinions of the type of Particle Swarm

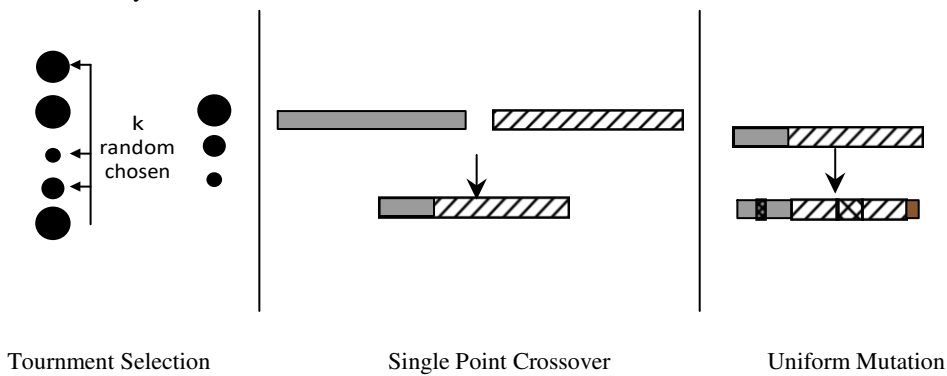

Tournment Selection          Single Point Crossover          Uniform Mutation

Fig. 1: Simple Genetic Algorithm operators

Optimization Algorithm the first one [5, 6] saw it as an evolutionary computation technique because it has evolutionary attributes:

- An initializing process that creates a population represents the random solution for each individual.
- Search for the best solution by creating a new better population.
- Every new population is based on the previous population.

On the other hand, the second opinion [7] views it as a stochastic optimization algorithm that shares similarities with other evolutionary algorithms.

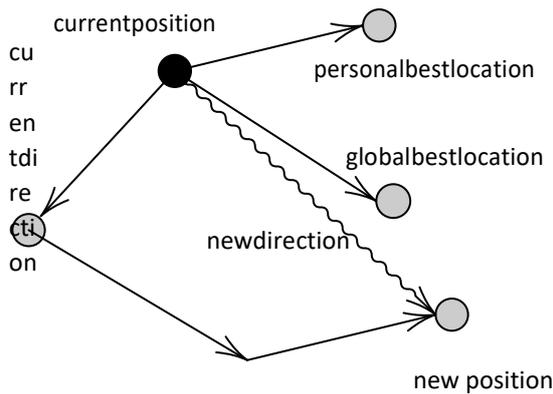

Fig. 2: PSO updating positions

The Algorithm is not complex and we can simplify the code as following:

Algorithm 2 PSO Algorithm

1: *initializeControllingParameters*()
2: *particles* ← *initializeFirstPopulation*()
3: while end condition is not satisfied do
4:     for each *particle* in *particles* do
5:         *calculateObjective*()
6:         *updateBestLocal*()
7:         *updateBestGlobal*()
8:     *updateInertiaWeight*()
9:     for each *particle* in *particles* do
10:        *updateVelocity*()
11:        *updatePosition*()
12: return *bestGlobalSolution*

Controlling parameters are:
- $N$: Size of the population.
- $c_1, c_2$: The weights which used in *updateVelocity*() that apply Eq. 3.2.
- $W_{min}, W_{max}$: The range of inertia weight ($\omega$) in Eq. 3.2 which *updateInertiaWeight*() work according to it.
- *Iter*: Max number of iterations allowed for the algorithm.

### C. Grey Wolf Optimization Algorithm

Grey wolf optimizer (GWO) is a meta-heuristic algorithm that is based on *Swarm Intelligence* and was first introduced [8] in 2014. Grey wolf optimizer algorithm represents a pack of grey wolves and it mimics its leadership pyramid and hunting behavior. The social hierarchy is simulated by categorizing the entire population of search-agents to four categories [9] based on their fitness, according to the objective task the more fit they are the higher rank they are in:

- Alpha: Have the authority to decide where to rest and when to hunt. They manage the group.
- Beta: Advisor to alpha and organizer for the pack. They are the best candidates to be an alpha in the case of death or retirement.
- Delta: They dominate omega and follow the orders of alpha and beta wolves.
- Omega: Lowest in rank. They follow other dominant wolves.

And in order to mimic the hunting behavior of the wolves they needed to mathematically model the phases known to us: *searching, encircling & harassing then attacking the prey*. We initially start all wolves at a random position then we evaluate them against the objective and the best 3 fitness produced are Alpha, Beta and delta respectively, the rest of the population are Omegas.

With randomly initialized variables $a$, $r_1$ and $r_2$ used to evaluate $A$ Eq. 3.3 and $C$ Eq. 3.4. And under the assumption of being the best then a prey (near-optimal solution) must be near them, the Alpha $\alpha$, Beta $\beta$ and Delta $\delta$ will re-position for next iteration in a radius defined by $a$ and $c$ around the current iteration position [8].

$$\tilde{A} = 2\sim a.r_1 - \sim a \qquad \text{(Eq. 3.3)}$$

$$\tilde{C} = 2r_2 \qquad \text{(Eq. 3.4)}$$

Now the rest of the population Omegas $\omega$ will re-position to a random point within a circle radius $r$ that is calculated from the positions of Alpha, Beta and Delta Eq. 3.5.

$$\vec{D}_{wolf} = \left| \vec{C}_n . \vec{X}_{wolf} - \vec{X}_\omega \right| \qquad \text{(Eq. 3.5)}$$

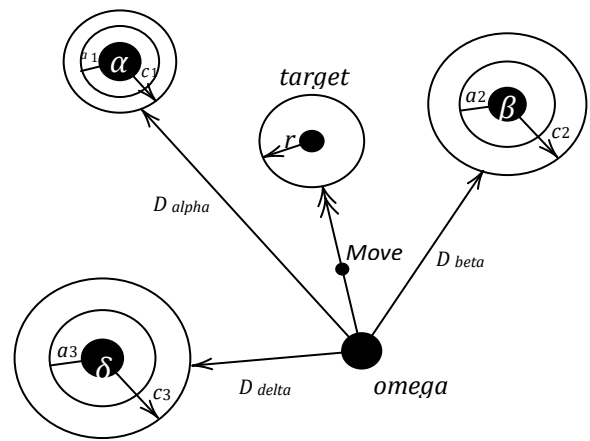

Fig. 3: GWO updating positions

*D. Differential Evolution Algorithm*

Differential Evolution was proposed by Rainer Strorn and Kenneth Price in 1997. [10] DE can be classified as an evolutionary optimization algorithm it has been widely used by researchers since its heuristic technique is independent of the domain of the problem it is used in many fields such as Computer Science and Engineering. it is a Stochastic population-based algorithm for solving single and multidimensional problems but does not follow the gradient if exist which means that it is not required to be differentiable.Differential Evolution optimizes the given problem by iteratively trying to enhance a population of genomes/chromosomes solutions and for each chromosome follows 3 operations mutation process followed by recombination and then selection and repeat the process till stopping criteria are admitted.

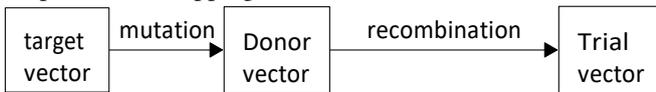

Selection of solutions is performed only after all trial vectors are computed then the greedy selection is performed between target and trial vectors.

- initialization A population of size *NP* where each individual is initialized randomly as follows

  $X_{i,j,G} = X_{i,jL} + rand_{i,j}[0,1] * (X_{i,jU} - X_{i,jL})$ (Eq. 3.6) $i \in \{1,2,3,...,N\}, j \in \{1,2,3,...,P\}$

  where $X^L$ is the lower bound of the variable $X_j$ and $X^U$ is the upper bound of the variable $X_j$

- Mutation

  The Donor vector ($V_i$) of a genome ($X_i$) is computed as

  $$\vec{V_i} = X_{r_1} + F(X_{r_2} + X_{r_2})$$ (Eq. 3.7)

  where F is a Scaling factor $F \in [0,2]$ and $X_r$ is a random vector where $i \in \{1,2,3,...,N_p\}$ and $r_1 \neq r_2 \neq r_3$, the target vector is not involved in mutation, this mutation is called self-referential mutation because the difference vector help in changing the search space. As the time passes, it alter the search space from exploration to exploitation.

- Recombination (Binomial Crossover) recombination is performed to increase the diversity of the population, components of the donor vector enter into the trial vector offspring vector in the following way, let $j_{rand}$ be a randomly chosen integer $\{1,2,...,D\}$.

  $u_{i,j,G} = \begin{cases} v_{i,j,G}, & \text{if } (rand_{i,j}[0,1]) \leq C_r \text{ or } i = i_{rand} \\ x_{i,j}, & \text{otherwise} \end{cases}$

- Selection "Survival of the fitter" principle in selection: the trial offspring vector is compared with the target vector (parent) and the one with high fitness value

  is approved to pass to next generation population.

  $X_{i,G+1} = \begin{cases} U_{i,G}, & \text{if } f(\tilde{U}_{i,G}) \leq f(\tilde{X}_{i,G}) \\ \vec{X}_{i,G}, & \text{otherwise} \end{cases}$

There are many mutation forms other than the one introduced in this paper. The algorithm can simply be as follows:

---
Algorithm 3 Differential algorithm
---
1: $Pop \leftarrow randomInitialize()$ . Eq. 3.6
2: while termination criteria not satisfied do
3:     for $i \leftarrow 1$ to *NP* do
4:         3 rand indexes $r_1, r_2, r_3$ where $r_1 \neq r_2 \neq r_3$
5:         $V_i \leftarrow mutation(r_1, r, r_3)$ . Eq. 3.7
6:         $j_{rand} \leftarrow rand(1,D)$
7:         for $k \leftarrow 1$ to *D* do
8:             if $rand(0,1) \leq C_r$ or $j = j_{rand}$ then
9:                 $U_{i,j,G} \leftarrow V_{i,j,G}$
10:         else
11:             $U_{i,j,G} \leftarrow X_{i,j,G}$
12:         $f(U_i^G) \leq f(X_i^G)$
13:         if $X^G + 1_i \leftarrow U_i^G$ then
14:
15:         else $X^G + 1_i \leftarrow X_i^G$
---

*E. Simulated Annealing*

A probabilistic technique used to find approximate optimum solution for nonlinear problems and non-convex functions,it is inspired by annealing process of metal, when metal is heated, the reshaping process be easier and the ability of ductility will increase, therefore the metal is bent more easily, with time the temperature decreases, therefore the ability of bending the metal decrease, The less temperature, the less flexibility, until the metal is forged and take its shape. The cooling process makes the heated metal be harder, overtime the ability to make a great change to the metal decreases, at the end the metal is forged.

Simulated annealing is discussed for the first time in Kirkpatrick, Gelatt, and Vecchi (1983) [11], it made very good results, this made many of researchers to improve simulation annealing algorithm and make applicable for more problems, it was proposed for solving problems that have a discrete search space as traveling salesman problem. However, there are several research papers introduced techniques to make it applicable to continuous domain problems. [12]

The idea of simulated annealing Algorithm is when the temperature (*T*) is large at first generations, therefore the algorithm will have a good chance to take the risk and go to the new point that is not good as the current point, this part of the algorithm makes it explore the dimension space, as the generations pass, the temperature (*T*) decreases and the probability of taking the risk decreases, from this moment it is just trying to improve and enhance the result and get sufficient point.

The algorithm of simulated annealing is as simple as following:

---

**Algorithm 4 Simulated Annealing**

1: $x_0 \leftarrow randomInitialize()$
2: $f_s \leftarrow f(x_0)$
3: for $i \leftarrow 1$ to $n$ do
4:     for $j \leftarrow 1$ to $m$ do
5:     $x_{new} \leftarrow nearpoint(x_0)$ 6: $\Delta E \leftarrow |f(x_{new}) - f_s|$
7:         if $f(x_{new}) < f_s$ then
8:             $x_0 \leftarrow x_{new}$
9:             $f_s \leftarrow f(x_{new})$
10:        else if $e^{-\Delta E/T} > rand(0,1)$ then
11:             $x_0 \leftarrow x_{new}$
12:             $f_s \leftarrow f(x_{new})$
13:     $T \leftarrow cooling(T)$

---

Variable of the algorithm as the following:

- $x_0$: Chosen individual in each generation
- $x_{new}$: neighbor point near $x_0$
- $f_s$: Fitness value
- $n$: Number of generations
- $m$: Number of neighbor points each generation
- $\Delta E$: Absolute difference of best fitness and new fitness value
- $T$: Temperature

Many research papers discuss how temperature ($T$) decreases, there are several methods and techniques for the cooling process. The cooling schedule is the process to decrease temperature ($T$) each generation, it is inefficient to make the temperature decrease by a large value in one generation that would make the algorithm stuck at an unsatisfying point with poor fitness value, and also decreasing by a small amount each generation is unreliable, it can make the algorithm accept new poor point at the last generation and this makes the algorithm fails to find an approximate solution, therefore the selection of appropriate cooling technique is important to success in finding the solution.

Logarithmic cooling, Geometrical cooling, Linear cooling, Adaptive and Arithmetic-Geometric cooling schedules [13] are cooling techniques, the fastest one of them is Geometrical cooling, its formula as following:

$$T_n = \alpha \cdot T_{n-1} \quad \text{(Eq. 3.8)}$$

Alpha factor ($\alpha$) is between 0 and 1 ($0 < \alpha < 1$), it is recommended that alpha ($\alpha$) be close to 1.

Is the algorithm guarantee finding the global optimum? the answer is no, sometimes it is stuck at a local minimum due to the randomness.

## IV. METHOD

There are well established meta-heuristic algorithms, we will only be discussing five evolutionary, swarm intelligence, and Stochastic Algorithms algorithms in this experiment which are Genetic Algorithm, Differential Evolution, Particle Swarm Optimization Algorithm, Grey Wolf Optimizer, and Simulated Annealing. the Comparison will be held between these algorithms.

We will test them against non-convex optimization problems which have a continuous domain, two optimization problems are frequently used for optimization algorithms tests [14], The first problem described as the following:

$$f(x,y) = 3(1-x^2)e^{-x^2-(y+1)^2} - \frac{e^{-(x+1)^2-y^2}}{3} - 10(\frac{x}{5} - x^3 - y^5)e^{-x^2-y^2}$$
(Eq. 4.1)

It is a simple function to test the performance of an optimization algorithm, It has two variables x and y, Domain = $\{[x,y]^T \in R^2 : -3 \leq x, y \leq 3\}$, It contains more than one local minimum, It is a good example to test the algorithms.

$$f(X) = 10n + \sum_{i=1}^{n}[x_i^2 - 10\cos(2\pi x_i)] \quad \text{(Eq. 4.2)}$$

Rastrigin function is well known for testing optimization algorithms, it has the ability to have a large number of dimensions, it is a robust function for testing optimization problem and it is difficult due to a large number of local minimum and large search space, Eq. 4.2 as shown, $n$ parameter represents the number of dimensions and $x$ describes a point in: $\{x : [x_1,...,x_n]^T \in R^n, -3 \leq x_i \leq 3$, where $1 \leq i \leq n\}$, and its global minimum at $x = (0,0)$, with objective $f(x) = 0$.

The goal of these non-convex functions is to find the global minimum or at least a near-optimal solution.

To compare these algorithms, we need to set standards to get a fair comparison, Python programming language was used as the implementation platform. The Aspects of this comparison is the fitness value at each generation and final fitness value.

## V. RESULTS AND DISCUSSION

We ran the tests on same machine and executed all algorithms with the same number of generations to ensure overall consistency of the results.

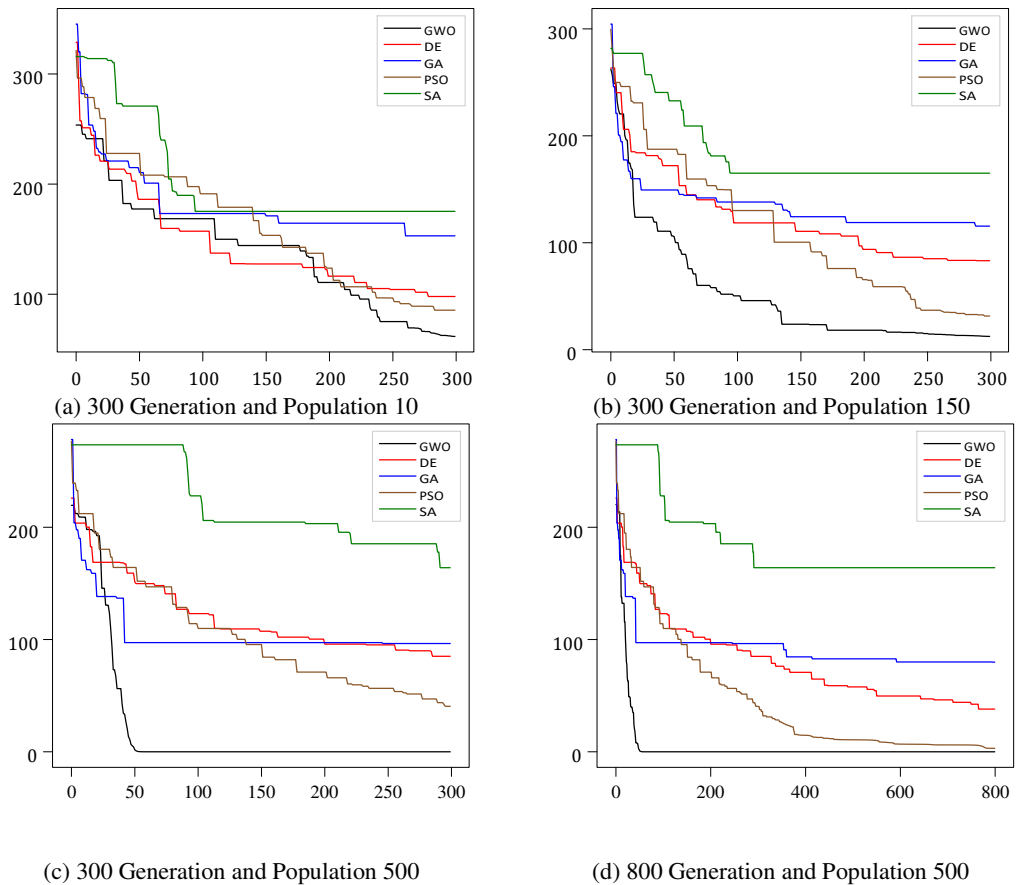

(a) 300 Generation and Population 10
(b) 300 Generation and Population 150
(c) 300 Generation and Population 500
(d) 800 Generation and Population 500

Fig. 4: Rastrigin objective function Eq. 4.2, 30 dimension search space

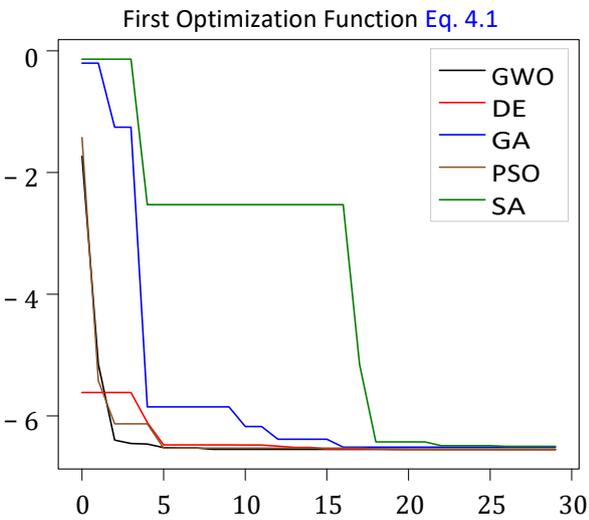

Fig. 5: Fitness Value and points at each generation

As shown in the figure above all the algorithms almost converge to the global minimum because of how simple the objective Eq. 4.1 is, but we can observe that particle swarm and grey wolf converges faster than other evolutionary algorithms because for particle swarm algorithm, The particles see each other state and decide to change their state to the best particle in their neighborhood also in grey wolf optimizer the pack updates its position based on the pack leaders position.

While genetic and differential algorithms genes do not have the ability to know the changes that had happened. [15]

For the second objective function with 30 dimensions Eq. 4.2 with $n = 10$, As shown in Figure 4, We can see that when the population was 10 all algorithms are almost enhance the best solution by the same rate in the first 100 generations, After that the difference between algorithms is started to be clear (Figure 4a). By increasing the size of population the difference between algorithms is increased (Figure 4b and Figure 4c) especially for gray wolf optimizer because implicitly assumes that the optimal solution is in the origin. Also, we can see *GA* and *SA* are trapped in local minimum but *GWO*, *PSO*, and *DE* are escaped from after number of generation, Figure 4d is the proof of this, Because when the number of generations increased to 800 the *PSO* and DE skipped from the local minimum which they trapped on them in Figure 4c which has 300 generations only.

## VI. CONCLUSION AND RECOMMENDATION

The effectiveness of meta-heuristic algorithms for constrained optimization problems is observed based on experimentation with benchmark functions. It is difficult to make a fair comparison because each algorithm has its own parameters, each parameter needs to be tuned but this tuning doesn't have a rule to follow.

But we can bring some observations from the demonstrated results. During the early generations, the only difference between algorithms is the randomness of the first population. Also, we observed that with an insufficient population no matter what algorithm you use you probably will end up with nearly the same result. Another observation would be that the higher your population is, the better your early generation converge to better fitness value.

Thought out the experiment we've noticed the *grey wolf optimizer* converging to the objective faster than expected with the population and generation increase Figure 4c. And we figured that the grey wolf optimizer algorithm implicitly assumes that the optimal solution is in the origin as mentioned before in Results and Discussion. Hence, with such inaccurate mathematical modeling we advise against using it on non-translated problems (*i.e.* optimal solution is at the origin).

There are some ideas that we may take into consideration in our future work such as parameter tuning for each meta-heuristic algorithm. Another important topic that worth investigating further is the reliability of the grey wolf algorithm when solving a problem where the origin point is the optimal solution, although it converges almost immediately, in general, its heuristic has no use in such problems and could be improved. also, we recommend making a comparison between Swarm Intelligence algorithms and Evolutionary Algorithms, it will be a valuable step to compare them with real-life optimization problems.